\pdfoutput=1
\documentclass{article}

\usepackage{times}
\usepackage{graphicx} 

\usepackage{latexsym}

\usepackage{algorithm}
\usepackage{algorithmic}

\usepackage{caption}
\usepackage{subcaption}
\usepackage{graphicx}
\usepackage{graphics}
\usepackage{amsmath}

\usepackage{authblk}

\usepackage{ijcai16}

\usepackage[utf8]{inputenc} 
\usepackage[T1]{fontenc}    
\usepackage{url}            
\usepackage{booktabs}       
\usepackage{amsfonts}       
\usepackage{nicefrac}       
\usepackage{microtype}      
\usepackage{url}

\usepackage{mathtools}
\usepackage{amsthm}

\title{Perceptual Reward Functions}

\author[Ashley Edwards et al.]
       {Ashley Edwards$^1$\thanks{This material is based upon work supported by the National Science Foundation Graduate Research Fellowship under Grant No. DGE-1148903 and the International Research Fellowship of the Japan Society for the Promotion of Science. }, Charles L. Isbell$^1$, and Atsuo Takanishi$^2$
       \\
       $^1$College of Computing, Georgia Institute of Technology, Atlanta,
       GA, USA\\
       aedwards8@gatech.edu, isbell@cc.gatech.edu\\
       $^2$Department of Modern Mechanical Engineering, Waseda University, Tokyo,
       Japan\\
       takanisi@waseda.jp\\
       }

\begin{document}
\maketitle
\begin{abstract}
Reinforcement learning problems are often described through rewards that indicate if an agent has completed some task. This specification can yield desirable behavior, however many problems are difficult to specify in this manner, as one often needs to know the proper configuration for the agent. When humans are learning to solve tasks, we often learn from visual instructions composed of images or videos. Such representations motivate our development of Perceptual Reward Functions, which provide a mechanism for creating visual task descriptions. We show that this approach allows an agent to learn from rewards that are based on raw pixels rather than internal parameters.
\end{abstract}

\section{Introduction}
Goals in reinforcement learning are often specified through rewards as a function of an agent's state variables. These variables have traditionally been tuned to the domain and include information such as the location of the agent or other objects in the world. The reward function then is inherently based on domain-specific representations. While such reward specifications can be sufficient enough to produce optimal behavior, more complex tasks might be difficult to express in this manner. Suppose a robot has a task of building origami figures. The environment would need to provide a reward each time the robot made a correct figure, thus requiring the program designer to~\emph{define} a notion of correctness for each desired configuration. Constructing a reward function for each model might become tedious and even difficult---what should the state variables for such a problem even be?

Raw pixels have recently become a popular state representation for reinforcement learning problems~\cite{mnih2015human}. With such inputs, we can abstract away the design of the relevant features of the agent's task. However, the rewards have often still been defined through parameters that are internal to the domain implementation. We aim to develop a general reward function based only on visual features that does not require manipulating such parameters when the task changes.

People often use visual sources to learn how to solve problems---be it from other people, diagrams, or videos. For example, we might look at an image of a completed origami figure to determine if our own model is correct. Our aim is to use similar visual descriptions for reinforcement learning tasks. We introduce Perceptual Reward Functions (PRFs), where the reward is based on how similar an agent's visual representation---such as an image from a camera or simulation---is to some goal representation. We will describe three approaches for representing tasks in this manner.

The rest of the paper is organized as follows. We begin by providing the background for our approach in Section~\ref{sec:background}. In Section~\ref{sec:approach} we introduce PRFs and describe approaches for representing tasks visually. We then provide empirical results in Section~\ref{sec:experiments}. In Section~\ref{sec:related} we discuss related work and in Section~\ref{sec:conclusion} we conclude.
\section{Background}
\label{sec:background}
\begin{figure}[h]
\centering
\begin{subfigure}{.32\columnwidth}
  \centering
  \includegraphics[width=.92\linewidth]{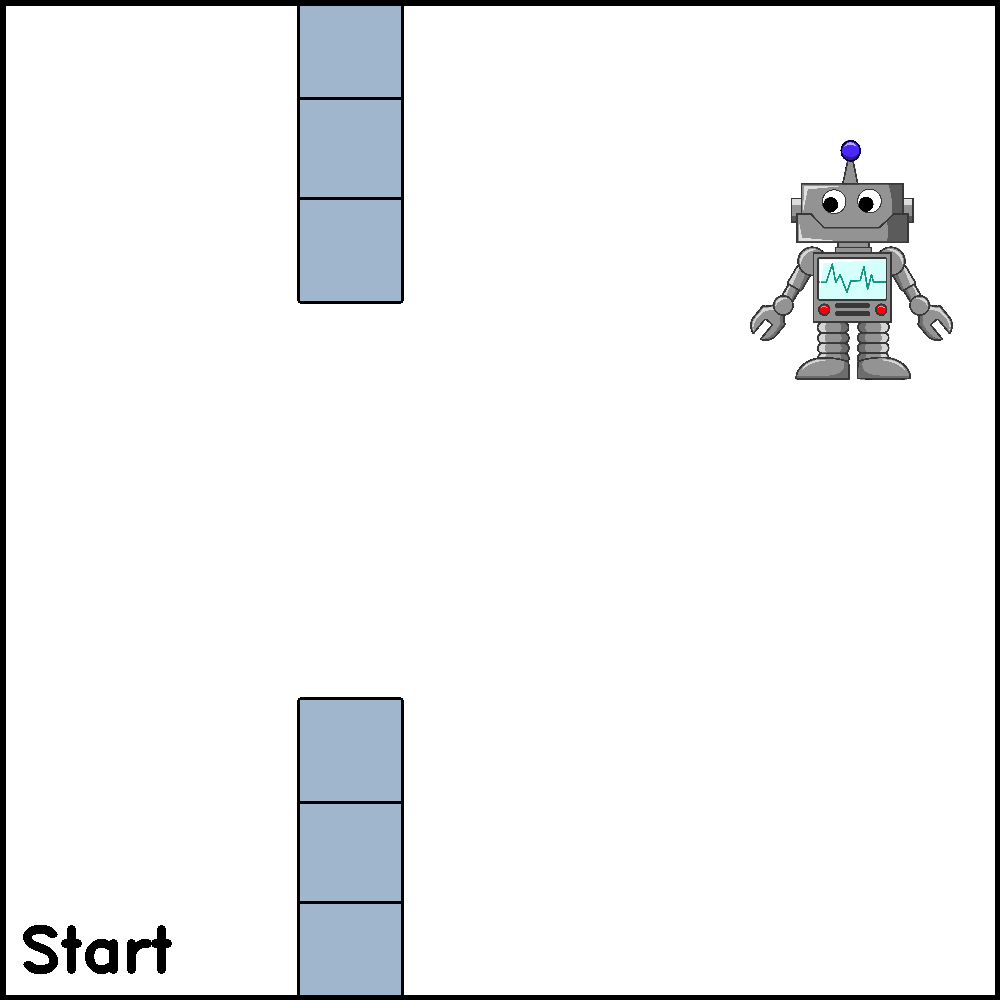}

\end{subfigure}%
\centering
\begin{subfigure}{.32\columnwidth}
  \centering
  \includegraphics[width=.92\linewidth]{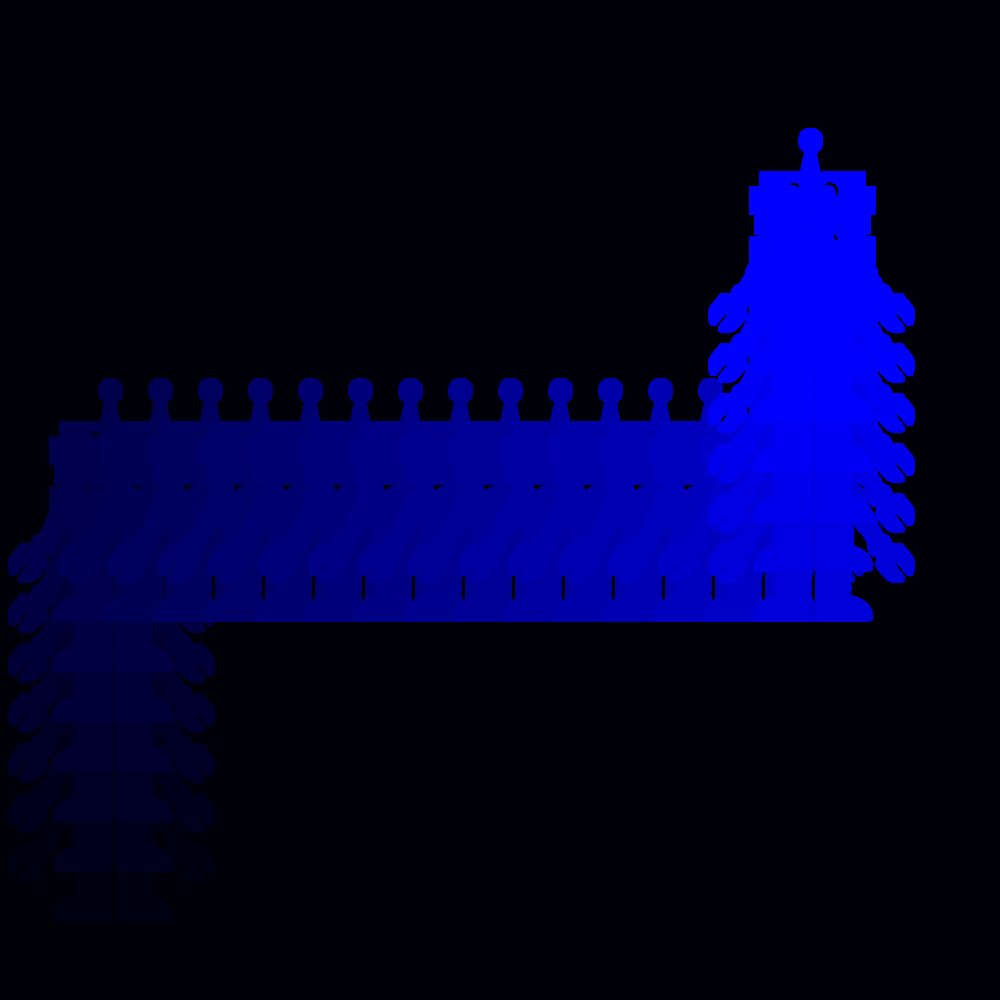}

\end{subfigure}%
\centering
\begin{subfigure}{.32\columnwidth}
  \centering
  \includegraphics[width=.92\linewidth]{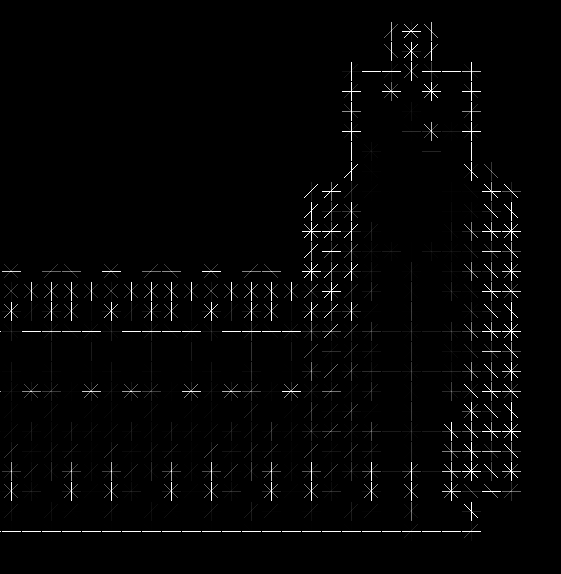}
\end{subfigure}%
\label{fig:hog}
\caption{The image on the left shows the result of an agent moving from the~\textbf{start} location to the top-right position. The middle image is the corresponding motion template. The image on the right is a zoomed-in visualization of the HOG features of the motion template.}
\label{fig:motionTemplate}
\end{figure}
We now provide a brief introduction to reinforcement learning and background for the visual state and goal representations that we introduce in Section~\ref{sec:motionTemplate}.
\subsection{Reinforcement learning}
Reinforcement learning problems are described through a Markov Decision Process $\langle S, A, P, R \rangle$~\cite{suttonbarto}. The set $S$ consists of states $s \in S$ that represent the current variables of an agent's environment. An agent takes actions $a \in A$ and receives rewards $r \in R(s)$ that depend on the current state. The transition function $P(s, a, s')$ represents the probability that the agent will land in state $s'$ after taking action $a$ in state $s$. The learning approach that we use is model-free and does not have access to $P$. A policy $\pi(s,a)$ represents the probability of taking action $a$ in state $s$.

Goals in reinforcement learning problems are often described solely through the reward function. An action-value, or Q-value, $Q(s, a)$ represents the expected discounted cumulative reward an agent will receive after taking action $a$ in state $s$, then following $\pi$ thereafter. We typically are interested in computing optimal Q-values:
$$
Q^*(s,a) = \max_{\pi} \mathbb{E}\Bigg[\sum_{k=0}^{\infty}\gamma^{k}r_{t+k+1} | s_t = s, a_t = a, \pi\Bigg]
$$
where $0 \le \gamma < 1$ is a discount factor that encodes how rewards retain their value over-time.

Q-values can be estimated through tabular methods that map every state and action to a value. However, with large or continuous state spaces, it is often necessary to compute the values with function approximation. In our case, the state inputs will be images and so we use a Deep Q-Network (DQN) to approximate the value function, as it has been empirically shown to perform well with visual inputs~\cite{mnih2015human}.
\subsection{Visual representations}
Our work compares an agent's visual representation to a visual goal representation. We now outline two techniques we use to ensure the representations are comparable.
\subsubsection{Motion templates}
A Motion Template (Figure~\ref{fig:motionTemplate}) is a 2D visual representation of motion that has occurred in a sequence of images---typically from the segmented frames of a video~\cite{davis1999recognizing,bobick2001recognition}. Movement that occurred more recently in time has a higher pixel intensity in the template than earlier motion and depicts both where and when motion occurred.

Calculating a motion template is an iterative process. The first step is to obtain a silhouette image of the motion that has occurred between each frame. The silhouette is computed by taking the absolute difference between two images and then computing the binary threshold, which sets all pixels below a threshold to $0$ and all pixels above the threshold to $1$.

A function $\Psi(\textbf{I})$ computes the motion template $\mu$ for a sequence of images $i_1, i_2, \dots, i_n \in \textbf{I}$. Let $\sigma_t$ represent a silhouette image at time $t$. To calculate the motion template $\mu$ of $\textbf{I}$, we first compute a silhouette image $\sigma_1, \sigma_2, \dots, \sigma_{n-1}$ between all consecutive images $(i_1, i_2), (i_2, i_3), \dots, (i_{n-1}, i_n)$. Then $\forall_{x,y}$, where $x$ and $y$ are respective column and row pixel locations, we can compute $\mu_{t,x,y}$ for time $t = 1, 2, \dots, n$:
\[
    \mu_{t,x,y} =
\begin{cases}
    \tau,& \text{if } \sigma_{t,x,y} > 0 \\
    0, & \text{else if }  \mu_{t-1, x, y} < (\tau - \delta) \\
    \mu_{t-1, x, y}, & \text{otherwise}
\end{cases}
\]
In words, the function increases the intensity of the pixel at $x, y$ if movement has occurred at the current iteration $t$. Here, $\delta$ and $\tau$ are both parameters that influence how much $\mu_t$ is decayed. The parameter $\tau$ is a representation for the current time in the sequence and increases as $t$ increases. The parameter $\delta$ represents the duration of the motion template and controls how quickly pixels decay. Essentially,  $\Psi(\textbf{I})$ layers the silhouette images and weights them by time.
\subsubsection{HOG features}
A Histogram of Oriented Gradients (HOG)~\cite{dalal2005} is a feature descriptor that is capable of representing information about local appearances and shapes within images. The first step for computing HOG features is to calculate the gradients of the image, which account for changes in intensities between adjacent pixels. Then, the image is divided into cells, which help describe local information.
The next step is to compute a histogram of the gradients for each cell. Finally, the histograms are concatenated into a feature vector that represents the HOG features. A visual representation of the features can be found in Figure~\ref{fig:motionTemplate}.
\section{Approach}
\label{sec:approach}
This work aims to provide a mechanism for describing goals without modifying internal reward values. Rather than stating: ``The task is complete when these specific configurations are met,'' visual task descriptions allow one to say: ``Here is how the completed task should~\emph{look}.'' Such an approach would be useful if the task requirements were difficult to program or if we did not have access to the agent's internal configuration, for example if we were an end-user teaching an agent to learn tasks in multiple environments. We now define some terminology for our approach and then describe three methods for visually describing tasks.
\subsection{Formalities}
We define a ~\textbf{Perceptual Template} as an image that is used to represent an agent's state or goal. We assume that the agent's state is a perceptual template that is derived either from a simulation or camera. We call this state the agent's~\textbf{Mirror state}.  We define a~\textbf{Goal Template}, $T_G$, as a perceptual template of the agent's goal. We define an~\textbf{Agent Template}, $T_A$, as a perceptual template that is comparable to $T_G$. We later describe three different representations for perceptual templates (Section~\ref{sec:representations}). First, however, we describe how to compute rewards given these templates.
\subsection{Perceptual reward functions}
A Perceptual Reward Function (PRF) computes a reward that represents how similar $T_A$ is to $T_G$. This is a general reward function that should remain unchanged across tasks---only the inputs to the PRF should vary, which is dissimilar to typical reward functions that require modifying domain-specific information when the task changes.

Formally, we can define a PRF as follows:
$$
F(T_A, T_G)  = \frac{1}{e^{D(T_A, T_G)}}
$$
where $F$ represents the perceptual reward function and $D$ is a distance metric. We use an exponential function to avoid dividing by $0$ and to separate rewards with similar distances.

The smallest distance that can be returned by $D$ is $0$, and so $F$ will return rewards that are less than or equal to $1$. As the distance between $T_A$ and $T_G$ increases, the output of $F$ will approach $0$. An optimal policy then should return actions that minimize the visual distance between $T_A$ and $T_G$. We now describe how $D$ can be computed.

There may be differences in both translation and scale between $T_A$ and $T_G$. To address these problems, we automatically crop the templates into the smallest axis-aligned rectangles surrounding each respective convex hull, where a set of points is represented by the non-black pixels in each template. If the template does not have a convex hull, for example if the image is black, then we do not crop the image. Next, we rescale either $T_A$ or $T_G$ so that it is as large as the other template. A function $H(T)$ computes this region for a perceptual template $T$ and then computes HOG features for it.

One reason to use the HOG descriptor is that the approach divides the image into cells, thus accounting for some differences in translation and scale. Additionally, the features give information about where movement should occur and so the agent should be motivated to take actions in the correct regions of its environment. Finally, the descriptor has no information about specific image features, such as color or texture.  Essentially, we can use HOG features to compare $T_A$ and $T_G$, even if they come some different sources.

For some tasks, we might wish to increase the number of cells used with HOG features for more acuity. However, increasing the number of cells also increases the time to compute the features. Thus, we keep the cell size as a parameter that can be input into the PRF. A simple approach is to set the cell size to be some fraction of the height, $h$, of the cropped region computed by $H$.

We now define the distance metric:
$$
D(T_A, T_G)  = \left\Vert{H(T_A) - H(T_G)}\right\Vert
$$
In words, we take the Euclidean distance between the HOG features of the cropped templates. We now discuss how one can obtain $T_A$ and $T_G$.
\subsection{Visual task descriptors}
\label{sec:representations}
One benefit of our approach is that it allows one to focus on task representation, rather than reward manipulation.  In order to create a PRF, one simply needs to define $T_A$ and $T_G$. We now outline three different task representations.
\subsubsection{Mirror task descriptors}
The first task representation we consider requires that $T_A$ be based on the agent's mirror state. We call this type of representation a~\textbf{Mirror Task Descriptor}. In the simplest form of a mirror task descriptor, $T_A$~\emph{is} the agent's mirror state. We call such a representation a~\textbf{Direct Task Descriptor}, as we can compare $T_G$ directly with the mirror state. This representation is convenient if we can represent how the agent's entire mirror state should look at the end of a task. For example, the image on the left of Figure~\ref{fig:motionTemplate} could be a direct task descriptor of where the robot should be located.

We might only know how to represent the relevant portions of a task. Suppose, for example, that we just know that the robot needs to be in the top right corner and do not know anything about the wall configurations. In this case, a~\textbf{Window Task Descriptor} would be appropriate. In a window representation, we represent $T_G$ as a cropped out window of the desired mirror state. In order to compute $T_A$, we use template matching~\cite{opencv} to find the maximally matching window within the agent's mirror state. The template matching algorithm slides $T_G$ across the mirror state and returns the window with the largest normalized correlation.
\subsubsection{Motion template task descriptor}
\label{sec:motionTemplate}
We now consider tasks that are based on a trajectory of motion. A~\textbf{Motion Template Task Descriptor} represents tasks using motion templates. As we noted in Section~\ref{sec:background}, a motion template is constructed from a sequence of images. We will soon describe how we can obtain two image sequences, $~\textbf{G}$ and $~\textbf{S}$, that will be used to compute $T_G$ and $T_A$, respectively.

Using motion templates for PRFs is appropriate when the goal can be represented through a trajectory of motion. One benefit of using a motion template for a PRF is that it does not track domain-specific features, thus allowing for task generality. In fact, motion templates contain~\emph{no} information about specific features, such as texture or color, thus allowing the source image sequence for $~\textbf{G}$ to be different than that of $~\textbf{S}_\textbf{t}$. For example, a person could create a sequence of images by recording herself performing the task, or a pre-existing image sequence, such as frames from a video, could be used. Depending on the agent representation, one might even be able to use a video of an animal or a cartoon character.

Now, we describe the simple process of obtaining the agent's image sequence $\textbf{S}$. At the beginning of each episode, we initialize the sequence such that $\textbf{S}_\textbf{0} = \{\}$. Each time the agent takes a step in an episode, its mirror state is added to $S$.

Given the goal and state sequences, we can compute their respective motion templates. That is, $T_G = \Psi(\textbf{G})$ and $T_A = \Psi(\textbf{S})$.
\subsection{State representation}
\label{sec:stateRepresentation}
\begin{figure}[htb]
\centering
\begin{subfigure}{.45\columnwidth}
  \centering
  \includegraphics[width=.8\linewidth]{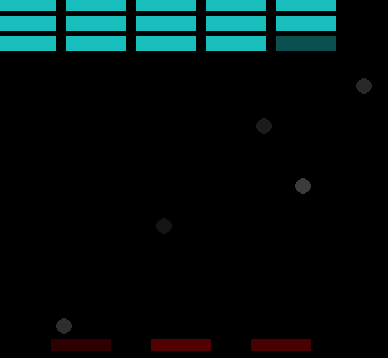}

    \label{fig:breakoutMem}
\end{subfigure}
\begin{subfigure}{.45\columnwidth}
  \centering
  \includegraphics[width=.8\linewidth]{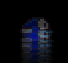}

    \label{fig:flappyMem}
\end{subfigure}
\caption{Zoomed in EMA states for Breakout and Flappy Bird with $\lambda=.7$. }
\label{stateRepresentations}
\end{figure}
We now briefly address the non-Markovian property of tasks with visual state inputs. Much of the information that is necessary to solve a specific task, such as velocity and relevant attributes of the environment, is lost when only a single image is used as input to a DQN. Traditionally, state inputs have been created from stacked frames in order to ensure the task remains Markovian~\cite{mnih2015human}. Additionally, recurrent neural networks have been used to learn the relevant history~\cite{hausknecht2015deep}. These approaches have been successful, but require more memory and introduce more parameters to be learned.

Our solution to this problem is simple. At each time-step, we take the Exponential Moving Average (EMA)~\cite{hunter1986exponentially}:
$$
a_t = (1-\lambda) s_t + \lambda a_{t-1}
$$
where $s_t$ is the agent's mirror state at time $t$ and $a_0 \coloneqq s_0$. The parameter $0 \le \lambda < 1$ represents how much memory should be saved overtime. The parameter essentially weights the past and the present separately. A value of $0$ would store no previous states.  We call $a_t$ an EMA state. This representation requires no more memory than a single mirror state. Additionally, the relevant history does not need to be learned since EMA states already~\emph{represent} the previously seen history. An example of an EMA state is show in Figure~\ref{stateRepresentations}.
\section{Experiments and results}
\label{sec:experiments}
\begin{figure}[htb]
\centering
\begin{subfigure}[t]{0.3\columnwidth}
  \centering
  \includegraphics[height=.7in]{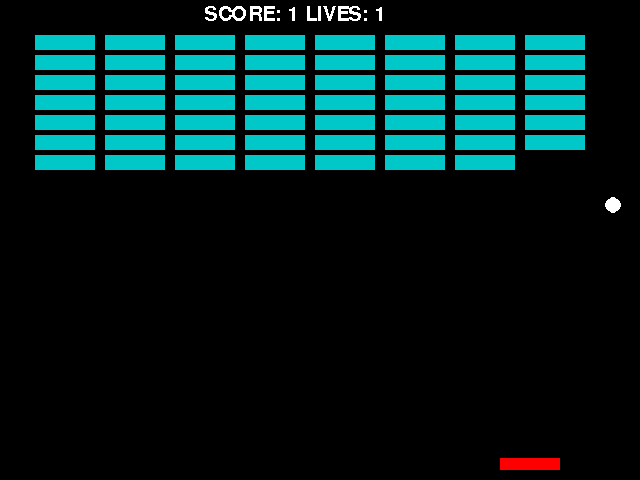}
  \caption{Breakout}
    \label{fig:breakout}
\end{subfigure}
\begin{subfigure}[t]{0.3\columnwidth}
  \centering
  \includegraphics[height=.7in]{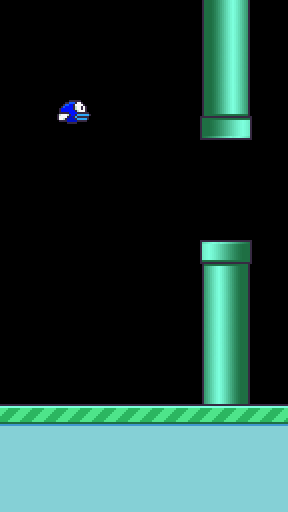}
  \caption{Flappy Bird}
    \label{fig:flappy}
\end{subfigure}
\begin{subfigure}[t]{0.3\columnwidth}
  \centering
  \includegraphics[height=.7in]{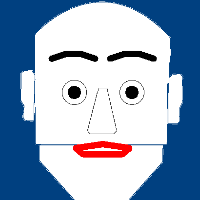}
  \caption{Kobian}
  \label{fig:kobian}
\end{subfigure}
\caption{Tasks used for evaluation. In Breakout, the agent must hit a pellet with a paddle to break all of the bricks in the game. In Flappy Bird, the agent must flap its wings to move itself between pipes. In the Kobian Simulator, the agent must move parts of its face to make expressions. Each of these images represent the agent's mirror state.}
\label{fig:domains}
\end{figure}
\begin{figure*}[htb]
\centering
\begin{subfigure}{.32\columnwidth}
  \centering
  \includegraphics[height=.8in]{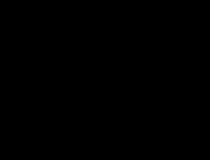}
  \caption{}
  \label{fig:atariTG}

\end{subfigure}
\begin{subfigure}{.26\columnwidth}
  \centering
  \includegraphics[height=.8in]{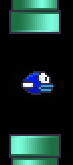}
  \caption{}
  \label{fig:flappyTG}

\end{subfigure}
\begin{subfigure}{.25\columnwidth}
  \centering
  \includegraphics[height=.8in]{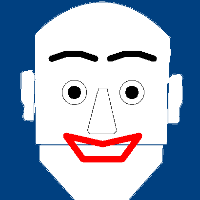}
    \caption{}
  \label{fig:happyTG}

\end{subfigure}
\centering
\begin{subfigure}{.25\columnwidth}
  \centering
  \includegraphics[height=.8in]{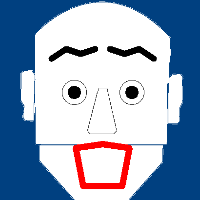}
    \caption{}
  \label{fig:surpriseTG}

\end{subfigure}
\begin{subfigure}{.25\columnwidth}
  \centering
  \includegraphics[height=.8in]{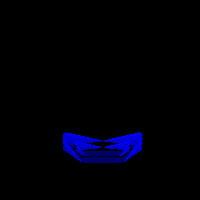}
    \caption{}

\end{subfigure}
\begin{subfigure}{.25\columnwidth}
  \centering
  \includegraphics[height=.8in]{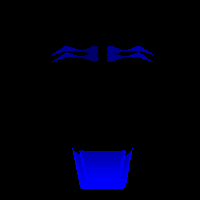}
  \caption{}

\end{subfigure}
\begin{subfigure}{.21\columnwidth}
  \centering
  \includegraphics[height=.8in]{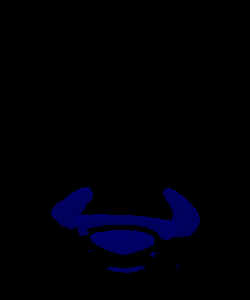}
  \caption{}
\end{subfigure}
\begin{subfigure}{.21\columnwidth}
  \centering
  \includegraphics[height=.8in]{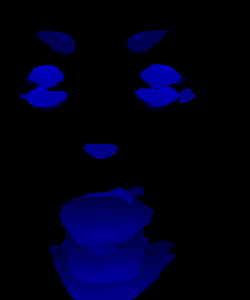}
  \caption{}
\end{subfigure}
\caption{Task Descriptors. From left to right: Breakout $T_G$, Flappy Bird $T_G$, Kobian Simulator Happy expression for VRF, Kobian Simulator Surprise Expression for VRF, KPRF Happy $T_G$, KPRF Surprise $T_G$, HPRF Happy $T_G$, HPRF Surprise $T_G$.}
\label{fig:taskRepresentations}
\end{figure*}
Our experiments aimed to show that PRFs are a general and feasible task representation for reinforcement learning. In particular, we aim to show that an agent is capable of learning from a visual task descriptor that did not require developing rewards that were based on task-specific features.

In each experiment, we evaluated a reward function based on the task's internal variables. We call this reward function a Variable Reward Function (VRF). We aim to show that a policy based on a PRF will yield behaviors that are at least as good as the behaviors learned with the VRF. We do not aim to show that one reward function allows the agent to learn faster than the other. Rather, we aim to show that we can represent tasks visually without needing to change the domain-specific parameters of the reward function.

We only allowed a limited number of steps for each task because in some tasks, the goal is to live forever, some tasks did not have a terminal function, and because the agent might never want to end a task because PRFs provide an infinite source of rewards, even if the agent is performing incorrectly. This is a problem that should be addressed in future work.

We used the same DQN architecture used in the state-of-the art for Deep Q-Learning~\cite{mnih2015human}, except the output layer depended on the number of actions for each domain and the input was a single image. We used an Adam Optimizer~\cite{kingma2014} for training. We followed an $\epsilon$-greedy approach while learning, with $\epsilon$ initialized to $1$ and decayed over time. The discount factor was set to $.99$. We evaluate the learned greedy policy after every $100$ episodes for each experiment. We now describe the experiments for three different task descriptors---a direct descriptor, a window descriptor, and a motion template descriptor.
\subsection{Mirror tasks}
\begin{figure}[htb]
\centering
\begin{subfigure}[t]{.4\linewidth}
  \centering
  \includegraphics[width=.99\linewidth]{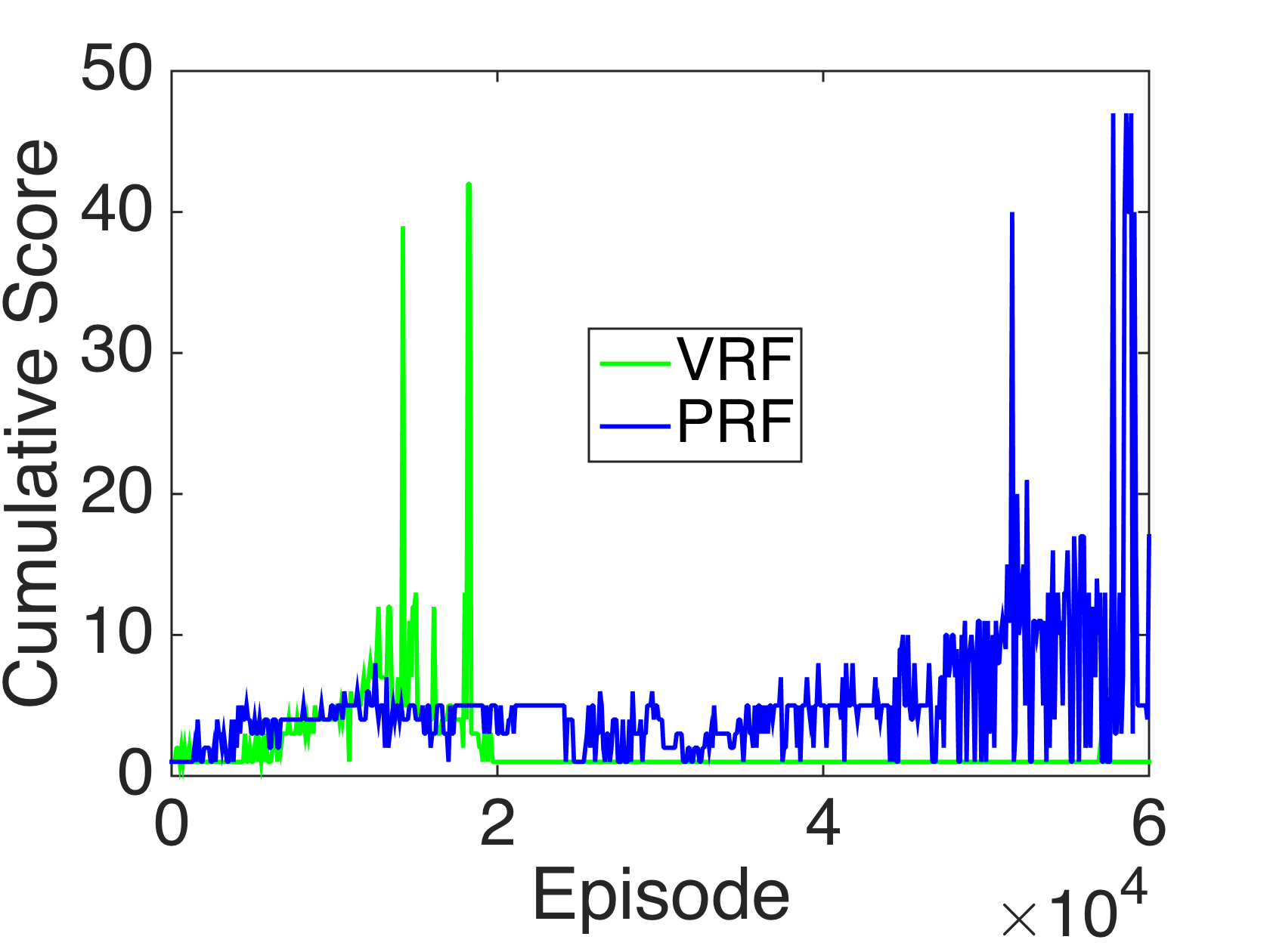}
  \caption{Breakout}
  \label{fig:atariResults}
\end{subfigure}
\begin{subfigure}[t]{.4\linewidth}
  \centering
  \includegraphics[width=.99\linewidth]{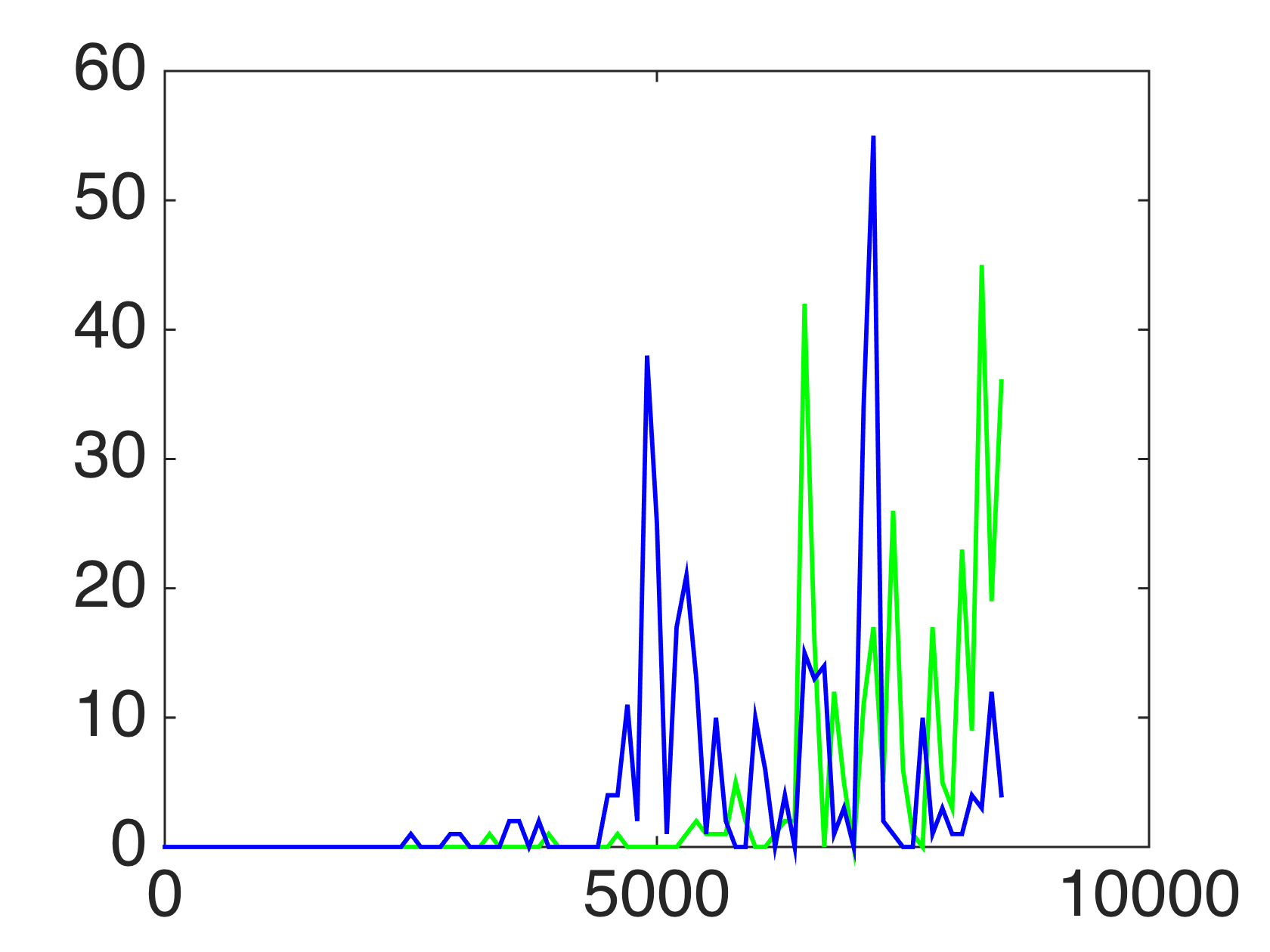}
  \caption{Flappy Bird}
  \label{fig:flappyResults}
\end{subfigure}
\caption{The results obtained in Breakout and Flappy Bird. We ran Breakout for 60,000 episodes and Flappy Bird for 8,500 episodes. In Breakout, the agent's score was incremented each time it hit a brick. In Flappy Bird, the agent's score was incremented when it moved through two pipes.}
\label{fig:results}
\end{figure}
\begin{figure}[htb]
\centering
\begin{subfigure}[t]{0.1\columnwidth}
  \centering
  \includegraphics[height=.7in]{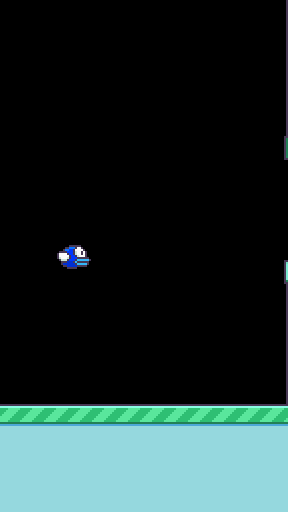}
\end{subfigure}%
\centering
\begin{subfigure}[t]{0.18\columnwidth}
  \centering
  \includegraphics[height=.7in]{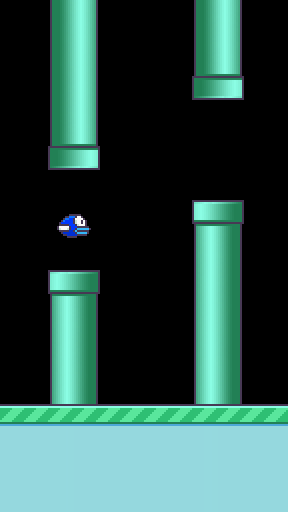}
\end{subfigure}%
\centering
\begin{subfigure}[t]{0.29\columnwidth}
  \centering
  \includegraphics[height=.7in]{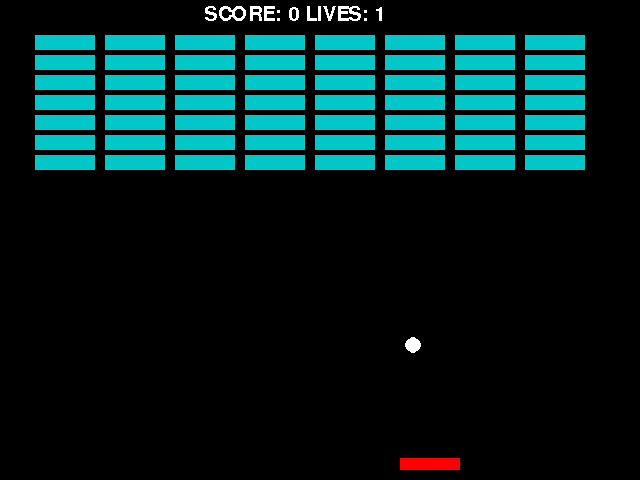}
\end{subfigure}%
\label{fig:hog}
\begin{subfigure}[t]{0.18\columnwidth}
  \centering
  \includegraphics[height=.7in]{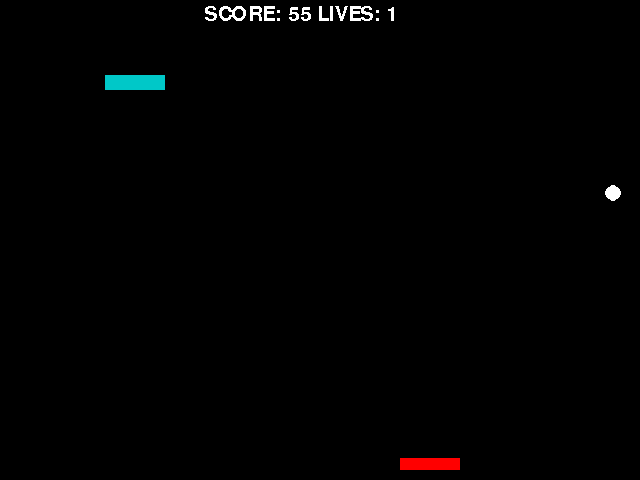}
\end{subfigure}%
\label{fig:hog}
\caption{The screens that yielded the lowest and highest rewards during a run of Flappy Bird and Breakout, respectively.}
\label{fig:topGames}
\end{figure}
\begin{figure*}[h]
\centering
\begin{subfigure}{.44\columnwidth}
  \centering
  \includegraphics[width=.96\linewidth]{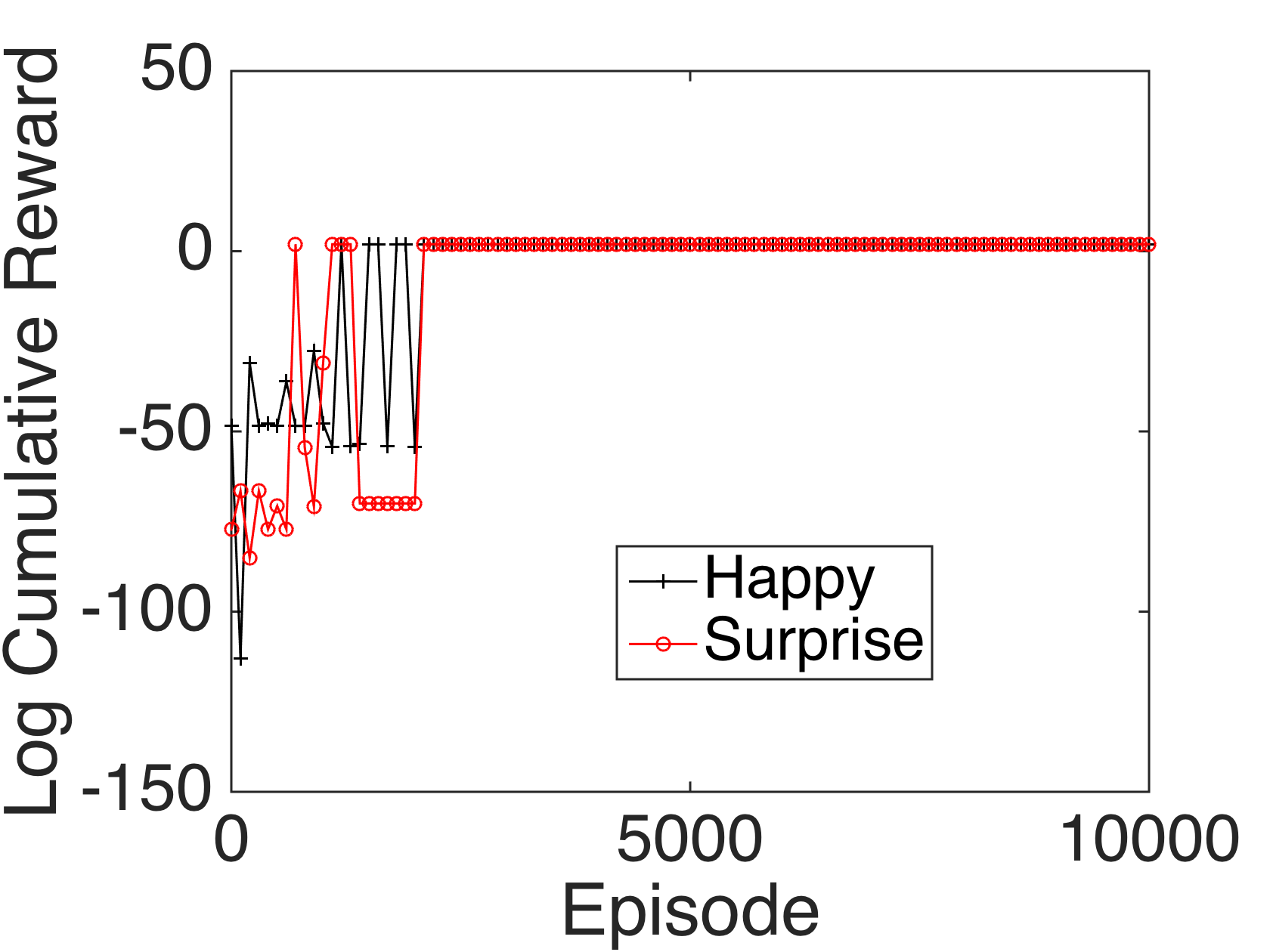}
  \caption{Kobian VRF}
\end{subfigure}
\begin{subfigure}{.4\columnwidth}
  \centering
  \includegraphics[width=.96\linewidth]{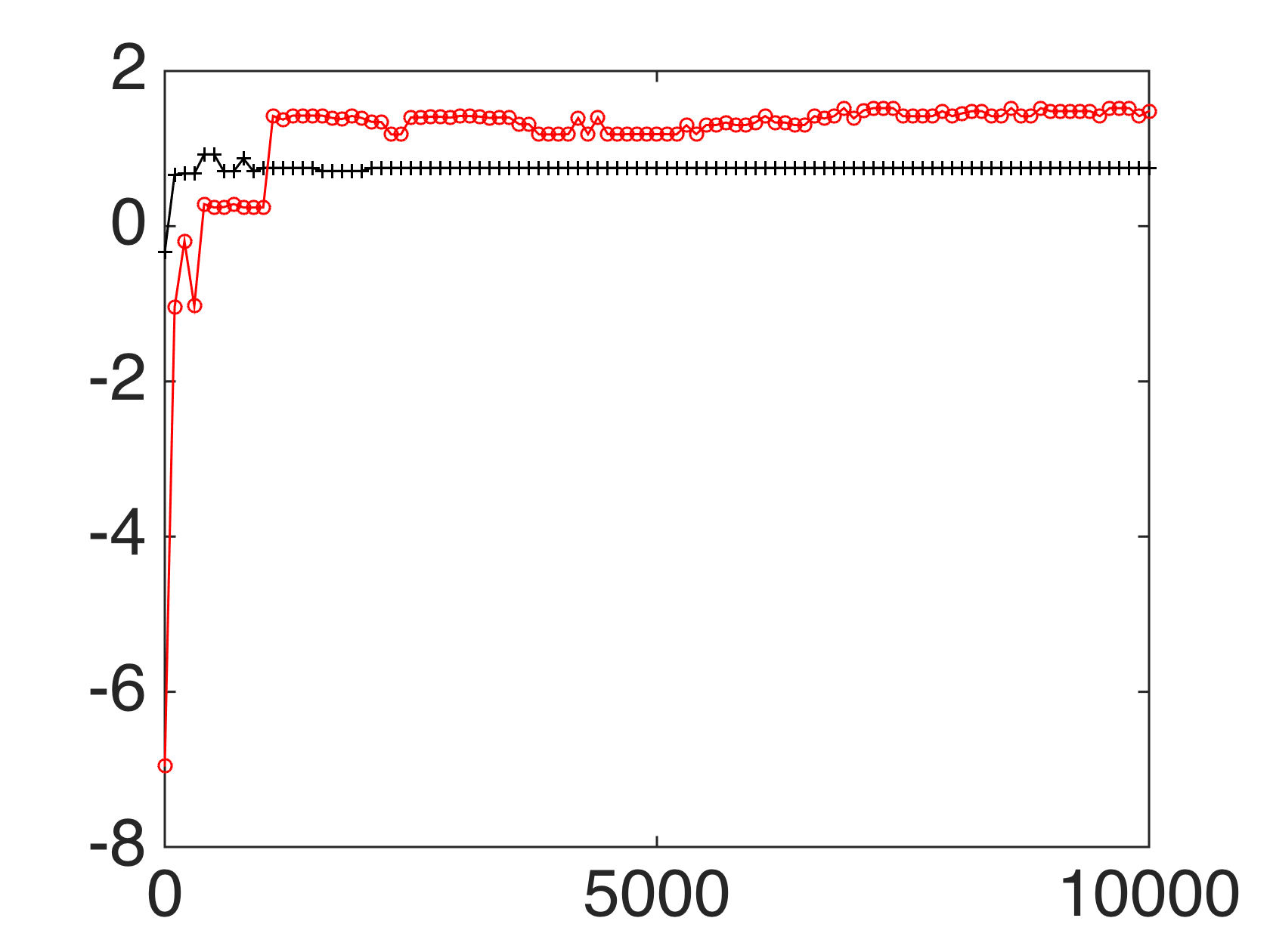}
  \label{fig:kprfResults}
  \caption{Kobian KPRF}
\end{subfigure}
\begin{subfigure}{.4\columnwidth}
  \centering
  \includegraphics[width=.96\linewidth]{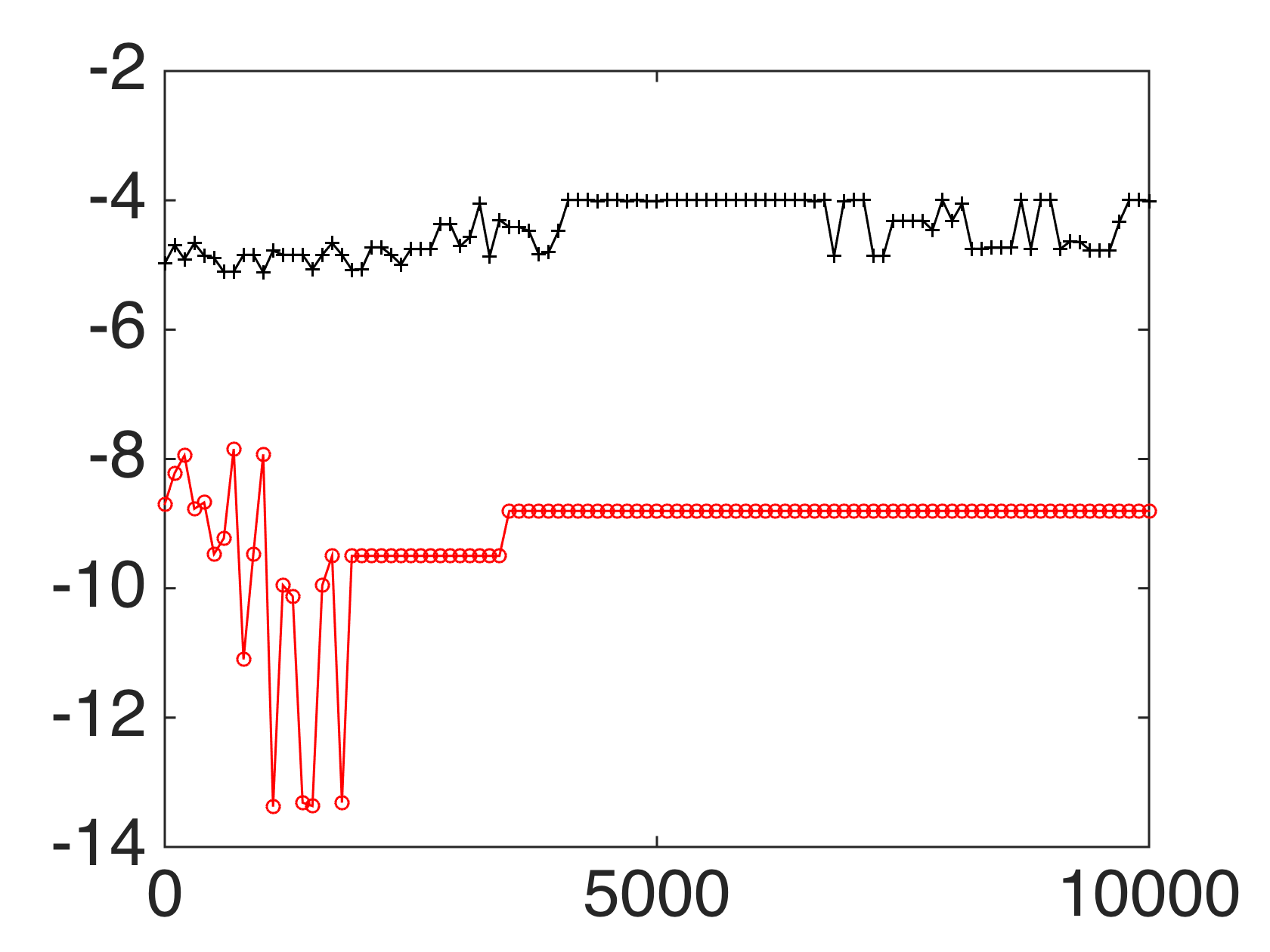}
  \label{fig:hprfResults}
  \caption{Kobian HPRF}
\end{subfigure}
\caption{The results obtained in the Kobian simulator. We ran the Kobian simulator for 10,000 episodes for each experiment.}
\label{fig:kobianResults}
\end{figure*}
In the mirror descriptor tasks, the state input for the DQN was the agent's EMA state with $\lambda = .7$. The agent's mirror state and $T_A$ were based on screens from the domain (Figure~\ref{fig:domains}). For computing the HOG features for the direct and mirror task descriptors, we used a cell size of $.03h$ and $.1h$, respectively. The initial learning rates were $.001$ and $.0001$.

We used the game Breakout (Figure~\ref{fig:breakout}) to evaluate a direct task descriptor. In this game, the agent essentially needs to take actions that make certain objects disappear, similar to Space Invaders and Tetris. Because of this, we represent $T_G$ as a black screen created with a simple paint software (Figure~\ref{fig:atariTG}). The agent had 3 actions to remain stationary or slide its pallet left or right. We limited the episode to 100 steps. We should note that for speed purposes, we did not use the common Arcade Learning Environment~\cite{bellemare2012arcade}.

We evaluated the PRF with this task representation and a VRF that returned a reward of $1$ for each brick hit after an action was taken. The results are shown in Figure~\ref{fig:atariResults}. The VRF agent does well initially, but eventually its performance drops. We likely need to tune the learning rate for the problem. Still, the agent was initially able to receive high score. The PRF agent was capable of achieving an even higher score that was close to optimal. This shows that a PRF based on a direct task descriptor is a feasible representation.

We used the game Flappy Bird (Figure~\ref{fig:flappy}) to evaluate a window task descriptor. The goal of Flappy Bird is to navigate a bird between pipes while it is moves horizontally across the screen. Therefore, we represent $T_G$ as an image of the bird between the pipes (Figure~\ref{fig:flappyTG}) that we obtained by simply opening a screen shot of the game in our paint software and dragging the bird. The agent had 2 actions to flap up and down. We limited the episode to 5,000 steps.

We evaluated the PRF with this descriptor and a VRF that gave a reward of 1 each time the agent was between the pipes, a reward of .1 each step, and -1 if it crashed. The results are shown in Figure~\ref{fig:flappyResults}. The PRF agent performed well and was able to achieve a higher score than VRF.  We have shown that the agent is capable of learning with a PRF based on a window task descriptor.
\subsubsection{Reward feasibility}
In both Flappy Bird and Breakout, the agent receives a reward from the PRF as long as it does not die. It is important then to evaluate if the rewards the agent is receiving are actually representative of $T_{A}$'s similarity to $T_G$. Even if the similarity metric were poor, the agent would still receive a reward and thus it would remain motivated to stay alive. Thus, we played each game and then obtained the mirror states that yielded the best and worst reward values, as shown in Figure~\ref{fig:topGames}. It is clear that relevant states are being rewarded appropriately.
\subsection{Motion template task}
\begin{figure}[htb]
\centering
\begin{subfigure}[t]{.15\columnwidth}
  \centering
  \includegraphics[width=.8\linewidth]{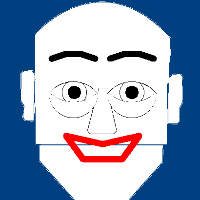}
   \vspace*{2mm}
\end{subfigure}
\begin{subfigure}[t]{.15\columnwidth}
  \centering
  \includegraphics[width=.8\linewidth]{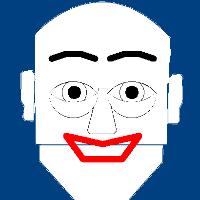}
   \vspace*{2mm}
\end{subfigure}
\begin{subfigure}[t]{.15\columnwidth}
  \centering
  \includegraphics[width=.8\linewidth]{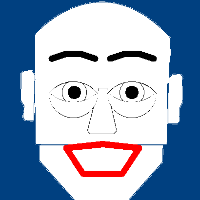}
   \vspace*{2mm}
\end{subfigure}
\begin{subfigure}[t]{.15\columnwidth}
  \centering
  \includegraphics[width=.8\linewidth]{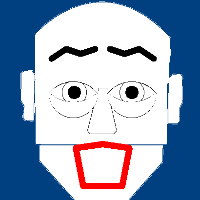}
\end{subfigure}
\begin{subfigure}[t]{.15\columnwidth}
  \centering
  \includegraphics[width=.8\linewidth]{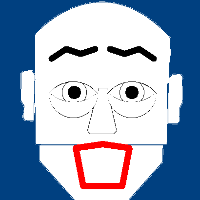}
\end{subfigure}
\begin{subfigure}[t]{.15\columnwidth}
  \centering
  \includegraphics[width=.8\linewidth]{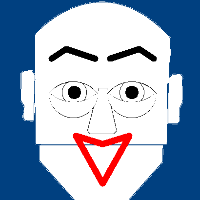}
\end{subfigure}

\caption{The faces that each experiment converged or nearly converged to. The first three images show the learned happy faces for the VRF, KPRF, and HPRF. The final three images show the learned surprise faces for the VRF, KPRF, and HPRF.}
\label{fig:bestFaces}

\end{figure}
In the motion template task, we used different DQN state inputs for the VRF and PRF. The input for the VRF was the agent's mirror state (Figure~\ref{fig:domains}) and the inputs for the PRF was the agent's motion template, or $T_A$. We did not use EMA states because motion templates already contain history and the VRF representation is Markovian. The cell size we used for the HOG features was $.05h$.

We used a simulation of the Kobian R-II's robot's~\cite{Kishi2012} face (Figure~\ref{fig:kobian}) to evaluate a motion template task representation. We initialized the learning rate to $.0001$. We aimed to train the agent to learn how to make facial expressions for happiness and surprise. The agent had 13 actions that allowed it to move different combinations of 24 vertices that made up the agent's eyebrows and mouth. There was not a terminal function that indicated when the goal is reached. Rather, we limited the agent's episode to 10 steps and allowed it to take a no-op action that did not effect its face or motion template, allowing the agent to decide when it had completed the task.

We used two PRFs for this task along with a VRF that gave a reward based on the distance between the vertices of the agent's eyebrows and mouth and those of a target face. For the first PRF, which we call the Kobian PRF (KPRF), we used the actions we took within the Kobian simulator to create the target face to generate the motion template for $T_G$. For the second PRF, which we call the Human PRF (HPRF), we used motion templates computed from videos from the Cohn-Kanade database~\cite{kanade2000comprehensive,lucey2010extended}, which consists of videos of humans making facial expressions. Each of the descriptors can be found in Figure~\ref{fig:taskRepresentations}.
We set $\tau$ to $.1$ and incremented by $.3$ for the HPRF and by $.4$ for the KPRF. We set $\delta$ to $\frac{|t + 1|}{4}$ for the HPRF and $\frac{|t + 1|}{3}$ for the KPRF, where $t$ was the current iteration of the motion template calculations, $\psi(\textbf{G})$ and $\psi(\textbf{S})$, respectively.

The results are shown in Figure~\ref{fig:results}. Both the VRF and the KPRF converged to the correct behavior. This shows that a motion template descriptor yields good results for a PRF when the agent's representation has the same source as the goal. The HPRF agent's performance for did not converge for the happy task, although the policy appeared to be approaching a specific set of actions. The agent did converge for the surprise task, but reward alone does not indicate if the agent was performing correctly. We can evaluate if the HPRF was learning the correct behavior by examining the faces it learned in Figure~\ref{fig:bestFaces}.

The agent learned similar faces as the KPRF and VRF agents. We did not expect the agent to learn the same expressions since its goal representation was based on a human and not the robot. Still, the agent learned to take the characteristic actions for each expression. For the happy expression, the agent stretched its mouth horizontally and opened it slightly. For the surprise expression, the agent raised its eyebrows and stretched its mouth vertically.

Overall, we have shown that motion template descriptors can be a good representation for PRFs, even when the source of $T_A$ is different than that of $T_G$.
\subsubsection{Reward feasibility}
\begin{figure}[htb]
\centering
\begin{subfigure}[t]{.4\linewidth}
  \centering
  \includegraphics[width=.99\linewidth]{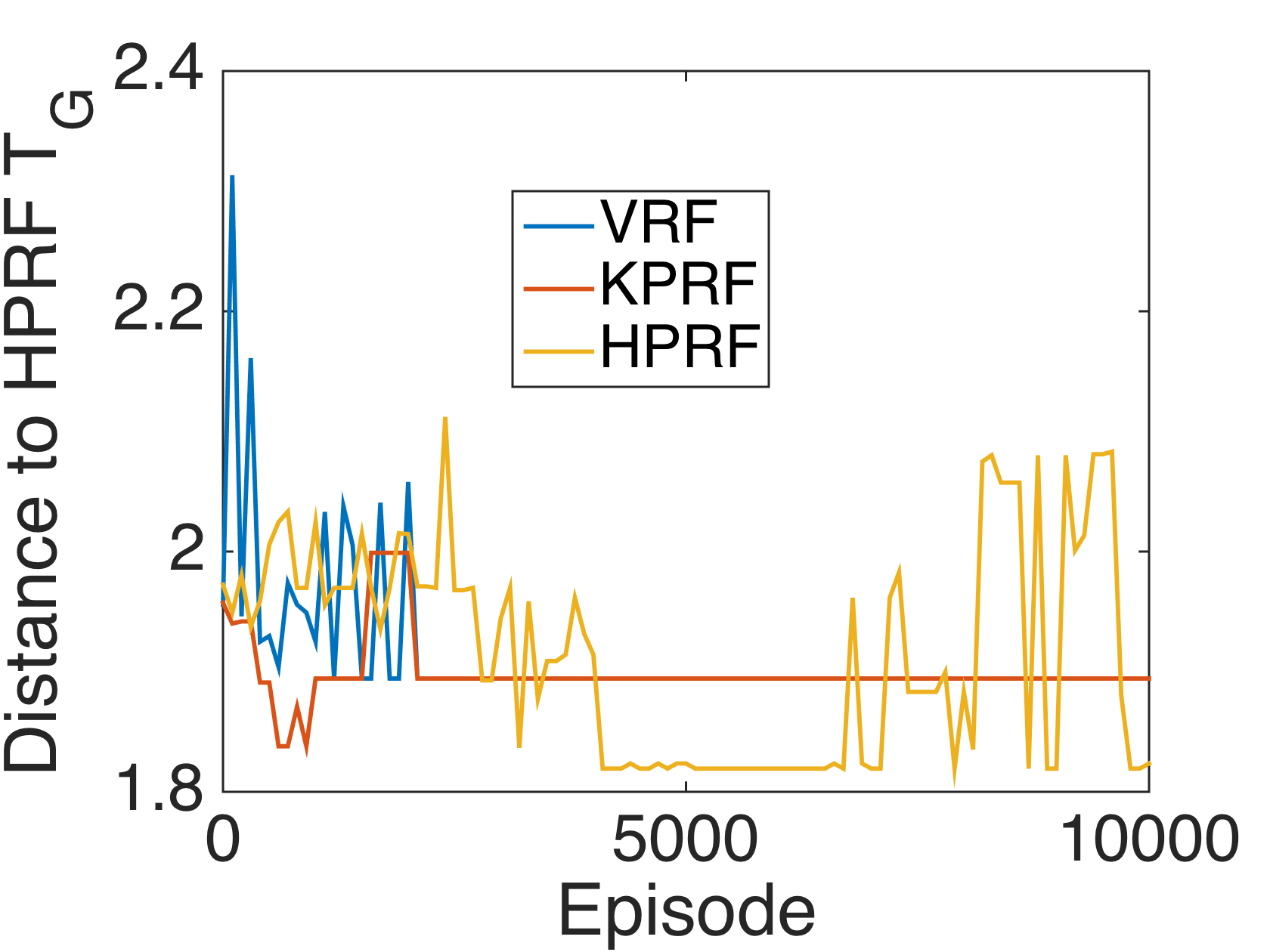}
  \caption{Happy}
  \label{fig:happinessDistance}
\end{subfigure}
\begin{subfigure}[t]{.4\linewidth}
  \centering
  \includegraphics[width=.99\linewidth]{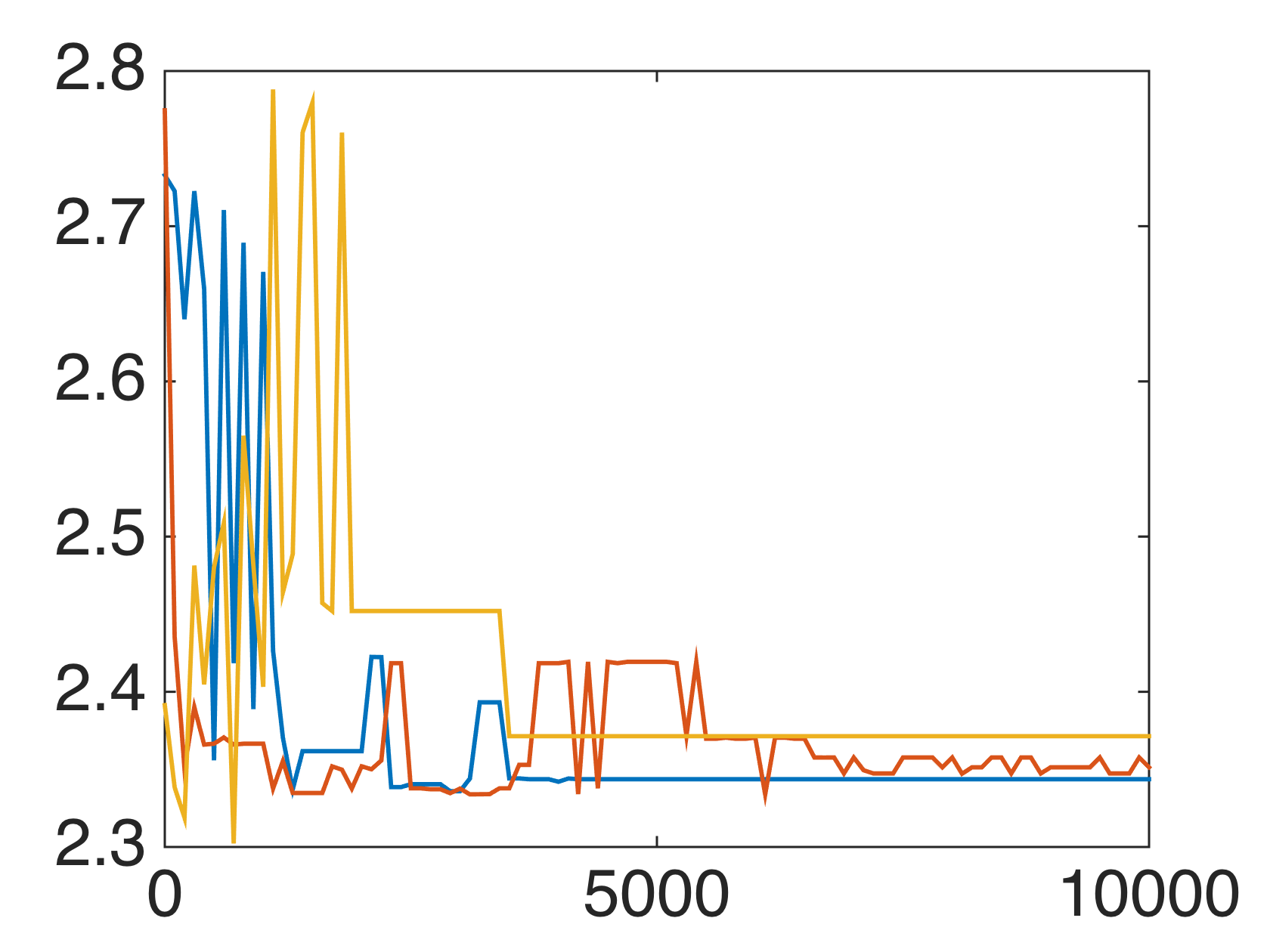}
  \caption{Surprise}
  \label{fig:surpriseDistance}
\end{subfigure}
\caption{The distance $D$ between the HPRF goal template, $T_G$, and the motion templates computed during each episode of the VRF, KPRF, and HPRF happiness and suprise tasks.}
\label{fig:distanceResults}
\end{figure}
Although we did not expect the HPRF agent to perform identically to the VRF and KPRF agents, the motion template for the HPRF should still represent the desired components of the task. We were interested in examining how similar the motion templates that resulted from the actions taken by the VRF and KPRF agents were to the $T_G$ representation of the HPRF. In particular, we measured the distance, $D$, between the two templates.

The results are shown in Figure~\ref{fig:distanceResults}. It is clear that the distance between the HPRF goal template and the motion templates computed for each reward function is decreasing. This shows that the motion templates represented the task well.
\section{Discussion and related work}
\label{sec:related}
We believe perceptual templates are a more natural form of expressing tasks for agents than traditional methods. Many works have aimed to simplify goal representations. For example, there has been work in representing goals through hand-drawn sketches~\cite{skubic2002hand,shah2012sketch}. Other works have focused on allowing goals to be represented through pre-existing sources. In one approach, a robot is trained to cook by showing it cooking videos~\cite{yang2015robot}. Specifically, the robot learned the necessary grasps and actions to take through classification. Another work uses information from the web to train robots~\cite{tenorth2011web}. In this approach, robots could learn how to solve problems by parsing the language in instruction websites such as~\url{wikihow.com} and~\url{ehow.com}.

Visual Servoing allows agents, particularly robots, to perform tasks that are represented through visual features~\cite{hutchinson1996tutorial}. While this approach is typically used for control, the idea of minimizing the error between a robot's camera image and some goal image is similar to our own work. Recently, we have seen a similar approach used with policy search that aims to minimize the difference between learned visual features from the agent's state space and a target representation~\cite{deepSpatial}. This work uses deep spatial autoencoders to learn important features in the agent's state space. Our work aims to develop a reward function that can be used for value-function approaches rather than policy search. Additionally, we were interested in examining how tasks can be represented even if the goal representation was not the same as the agent's. In this case, learned features points about the agent's state space might not map to those of the goal.

Still, our work may appear to be moving in the opposite direction of the field of reinforcement learning. The purpose of using deep learning as function approximator for Q-learning is indeed to avoid developing hand-crafted features. Although we use visual state inputs, the features used by the reward function are not learned. Additionally, we do need to set quite a few parameters for each task representation. However, HOG features are general enough to be used across multiple tasks, and the parameters used in our approach can be thought of as any other learning parameters that might need to be tuned. More to the point, we have not yet seen an approach where the reward function can be specified without knowledge of internal~\emph{task} parameters. With our approach, one can avoid developing hand-crafted reward functions and focus on how the task can be represented visually.
\section{Conclusion}
\label{sec:conclusion}
In this work, we have shown that PRFs allow one to create visual task descriptions without modifying domain-specific reward parameters. We introduced three different task representations and have shown empirically that they can yield desirable behavior. In conclusion, we have shown that PRFs are a feasible and general approach for representing tasks.
\bibliographystyle{named}
\bibliography{ijcai16}

\begin{thebibliography}{}

\bibitem[\protect\citeauthoryear{Bellemare \bgroup \em et al.\egroup
  }{2012}]{bellemare2012arcade}
Marc~G Bellemare, Yavar Naddaf, Joel Veness, and Michael Bowling.
\newblock The arcade learning environment: An evaluation platform for general
  agents.
\newblock {\em Journal of Artificial Intelligence Research}, 2012.

\bibitem[\protect\citeauthoryear{Bobick and
  Davis}{2001}]{bobick2001recognition}
Aaron~F Bobick and James~W Davis.
\newblock The recognition of human movement using temporal templates.
\newblock {\em Pattern Analysis and Machine Intelligence, IEEE Transactions
  on}, 23(3):257--267, 2001.

\bibitem[\protect\citeauthoryear{Bradski and Kaehler}{2008}]{opencv}
Gary Bradski and Adrian Kaehler.
\newblock {\em Learning OpenCV: Computer vision with the OpenCV library}.
\newblock " O'Reilly Media, Inc.", 2008.

\bibitem[\protect\citeauthoryear{Dalal and Triggs}{2005}]{dalal2005}
Navneet Dalal and Bill Triggs.
\newblock Histograms of oriented gradients for human detection.
\newblock In {\em Computer Vision and Pattern Recognition, 2005. CVPR 2005.
  IEEE Computer Society Conference on}, volume~1, pages 886--893. IEEE, 2005.

\bibitem[\protect\citeauthoryear{Davis}{1999}]{davis1999recognizing}
James Davis.
\newblock Recognizing movement using motion histograms.
\newblock {\em Technical Report 487, MIT Media Lab}, 1(487):1, 1999.

\bibitem[\protect\citeauthoryear{Finn \bgroup \em et al.\egroup
  }{2015}]{deepSpatial}
Chelsea Finn, Xin~Yu Tan, Yan Duan, Trevor Darrell, Sergey Levine, and Pieter
  Abbeel.
\newblock Deep spatial autoencoders for visuomotor learning.
\newblock {\em reconstruction}, 117(117):240, 2015.

\bibitem[\protect\citeauthoryear{Hausknecht and
  Stone}{2015}]{hausknecht2015deep}
Matthew Hausknecht and Peter Stone.
\newblock Deep recurrent q-learning for partially observable mdps.
\newblock {\em arXiv preprint arXiv:1507.06527}, 2015.

\bibitem[\protect\citeauthoryear{Hunter}{1986}]{hunter1986exponentially}
J~Stuart Hunter.
\newblock The exponentially weighted moving average.
\newblock {\em J. Quality Technol.}, 18(4):203--210, 1986.

\bibitem[\protect\citeauthoryear{Hutchinson \bgroup \em et al.\egroup
  }{1996}]{hutchinson1996tutorial}
Seth Hutchinson, Gregory~D Hager, Peter Corke, et~al.
\newblock A tutorial on visual servo control.
\newblock {\em Robotics and Automation, IEEE Transactions on}, 12(5):651--670,
  1996.

\bibitem[\protect\citeauthoryear{Kanade \bgroup \em et al.\egroup
  }{2000}]{kanade2000comprehensive}
Takeo Kanade, Jeffrey~F Cohn, and Yingli Tian.
\newblock Comprehensive database for facial expression analysis.
\newblock In {\em Automatic Face and Gesture Recognition, 2000. Proceedings.
  Fourth IEEE International Conference on}, pages 46--53. IEEE, 2000.

\bibitem[\protect\citeauthoryear{Kingma and Ba}{2014}]{kingma2014}
Diederik Kingma and Jimmy Ba.
\newblock Adam: A method for stochastic optimization.
\newblock {\em arXiv preprint arXiv:1412.6980}, 2014.

\bibitem[\protect\citeauthoryear{Kishi \bgroup \em et al.\egroup
  }{2012}]{Kishi2012}
Tatsuhiro Kishi, Takuya Otani, Nobutsuna Endo, Przemyslaw Kryczka, Kenji
  Hashimoto, K~Nakata, and Atsuo Takanishi.
\newblock Development of expressive robotic head for bipedal humanoid robot.
\newblock In {\em Intelligent Robots and Systems (IROS), 2012 IEEE/RSJ
  International Conference on}, pages 4584--4589. IEEE, 2012.

\bibitem[\protect\citeauthoryear{Lucey \bgroup \em et al.\egroup
  }{2010}]{lucey2010extended}
Patrick Lucey, Jeffrey~F Cohn, Takeo Kanade, Jason Saragih, Zara Ambadar, and
  Iain Matthews.
\newblock The extended cohn-kanade dataset (ck+): A complete dataset for action
  unit and emotion-specified expression.
\newblock In {\em Computer Vision and Pattern Recognition Workshops (CVPRW),
  2010 IEEE Computer Society Conference on}, pages 94--101. IEEE, 2010.

\bibitem[\protect\citeauthoryear{Mnih \bgroup \em et al.\egroup
  }{2015}]{mnih2015human}
Volodymyr Mnih, Koray Kavukcuoglu, David Silver, Andrei~A Rusu, Joel Veness,
  Marc~G Bellemare, Alex Graves, Martin Riedmiller, Andreas~K Fidjeland, Georg
  Ostrovski, et~al.
\newblock Human-level control through deep reinforcement learning.
\newblock {\em Nature}, 518(7540):529--533, 2015.

\bibitem[\protect\citeauthoryear{Shah \bgroup \em et al.\egroup
  }{2012}]{shah2012sketch}
Danelle Shah, Joseph Schneider, and Mark Campbell.
\newblock A sketch interface for robust and natural robot control.
\newblock {\em Proceedings of the IEEE}, 100(3):604--622, 2012.

\bibitem[\protect\citeauthoryear{Skubic \bgroup \em et al.\egroup
  }{2002}]{skubic2002hand}
Marjorie Skubic, Sam Blisard, Andy Carle, and Pascal Matsakis.
\newblock Hand-drawn maps for robot navigation.
\newblock In {\em AAAI Spring Symposium, Sketch Understanding Session},
  page~23, 2002.

\bibitem[\protect\citeauthoryear{Sutton and Barto}{1998}]{suttonbarto}
Richard~S Sutton and Andrew~G Barto.
\newblock {\em Reinforcement learning: An introduction}, volume~1.
\newblock Cambridge Univ Press, 1998.

\bibitem[\protect\citeauthoryear{Tenorth \bgroup \em et al.\egroup
  }{2011}]{tenorth2011web}
Moritz Tenorth, Ulrich Klank, Dejan Pangercic, and Michael Beetz.
\newblock Web-enabled robots.
\newblock {\em Robotics \& Automation Magazine, IEEE}, 18(2):58--68, 2011.

\bibitem[\protect\citeauthoryear{Yang \bgroup \em et al.\egroup
  }{2015}]{yang2015robot}
Yezhou Yang, Yi~Li, Cornelia Fermuller, and Yiannis Aloimonos.
\newblock Robot learning manipulation action plans by {“Watching”}
  unconstrained videos from world wide web.
\newblock In {\em {AAAI} 2015}, Austin, {US}, jan 2015.

\end{thebibliography}

\end{document}